\pdfoutput=1
\documentclass[a4paper,fleqn]{cas-dc}
\usepackage{graphicx}
\usepackage[numbers,square]{natbib}
\usepackage{float}

\def\tsc#1{\csdef{#1}{\textsc{\lowercase{#1}}\xspace}}
\tsc{WGM}
\tsc{QE}

\begin{document}
\let\WriteBookmarks\relax
\def\floatpagepagefraction{1}
\def\textpagefraction{.001}

\shorttitle{}    

\shortauthors{}  

\title [mode = title]{VerteNet - A Multi-Context Hybrid CNN Transformer for Accurate Vertebral Landmark Localization in Lateral Spine DXA Images} 

\author[1,2]{Arooba Maqsood}
\ead{a.maqsood@ecu.edu.au}

\author[1,2]{Zaid Ilyas}
\author[1,2]{Afsah Saleem}
\author[3]{Erchuan Zhang}
\author[1,2]{David Suter}
\author[7]{Parminder Raina}
\author[2,4]{Jonathan M. Hodgson}
\author[5]{John T. Schousboe}
\author[6]{William D. Leslie}
\author[1,2]{Joshua R. Lewis}
\author[1,2,8]{Syed Zulqarnain Gilani}

\affiliation[1]{organization={Center for AI \& ML, School of Science, Edith Cowan University}, 
            city={Joondalup},
            country={Western Australia}}
\affiliation[2]{organization={Nutrition and Health Innovation Research Institute, Edith Cowan University},
            city={Joondalup}, 
            country={Western Australia}}
\affiliation[3]{organization={School of Science, Sun Yat-sen University},
            city={Guangzhou},
            country={China}}
\affiliation[4]{organization={School of Medical and Health Sciences, Edith Cowan University},
            city={Joondalup},
            country={Western Australia}}
\affiliation[5]{organization={Park Nicollet Clinic and HealthPartners Institute},
            state={Minnesota},
            country={USA}}
\affiliation[6]{organization={Department of Medicine and Radiology, University of Manitoba},
            state={Manitoba},
            country={Canada}}
\affiliation[7]{organization={Department of Health Sciences, McMaster University}, 
            city={Hamilton},
            country={Canada}}
\affiliation[8]{organization={Computer Science and Software Engineering, The University of Western Australia},
            city={Perth},
            country={Western Australia},}

\cortext[1]{Corresponding Author: Arooba Maqsood}

\begin{abstract}
Vertebral Landmarks Localization (VLL) in Dual Energy X-Ray Absorptiometry (DXA)-based Lateral Spine Imaging (LSI) plays a critical role in evaluating spinal alignment, Vertebral Fracture Assessment (VFA), and facilitating intervertebral guide placement for Abdominal Aortic Calcification (AAC) quantification. While DXA LSI offers advantages such as reduced cost and lower radiation exposure, its analysis remains challenging due to a low signal-to-noise ratio and imaging artifacts. Artificial Intelligence (AI) presents a promising avenue for improving the precision and accuracy of VLL. This study introduces a novel architecture that employs dual-resolution attention mechanisms to capture both fine-grained local details and broader contextual information. Our approach enhances feature integration by leveraging skip connections and decoder layers through dual-resolution self- and cross-attentions. This design improves the model’s ability to learn complex patterns, ensuring precise vertebral corner localization while maintaining both local and global contextual awareness. We evaluated our proposed framework on DXA LSIs acquired from various machines and found that it outperformed recent state-of-the-art architectures, trained for VLL, achieving a normalized mean error of 4.92 and a normalized median error of 2.35. The proposed framework, VerteNet, enables highly accurate VLL in DXA LSI images from diverse machines and demonstrates superior robustness to low signal-to-noise ratios, owing to its enhanced ability to capture both fine-grained local details and broader contextual information.
\end{abstract}


\maketitle

\section{Introduction}\label{introduction}
\begin{figure*}
\centerline{\includegraphics[width=0.89\textwidth]{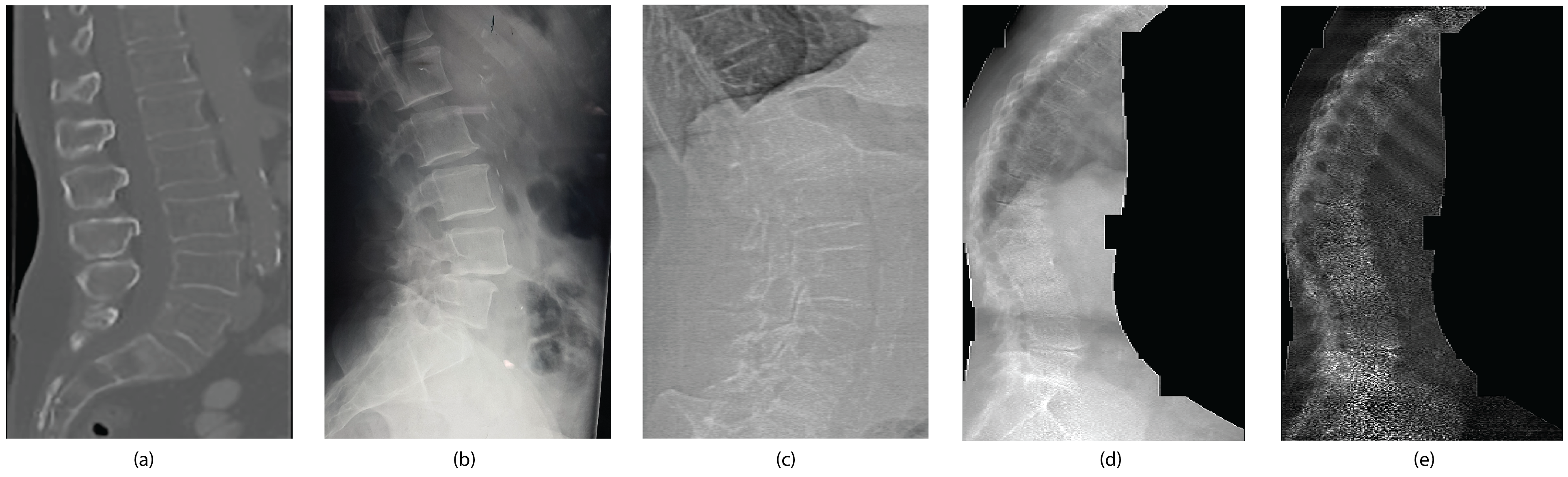}}
\caption{(a) CT Lateral Spine Imaging – the gold standard, slower with highest radiation exposure \cite{ct}. (b) Digital X-Ray Imaging – a faster option with lower radiation exposure than CT \cite{dxi}. (c) Hologic DXA SE variant – quickest with lowest radiation exposure, though susceptible to artifacts such as bowel gas. (d) and (e) GE DXA SE and DE variants – equipped with radiation-reducing technology (black regions), offering comparatively fast imaging with low radiation exposure.}
\label{fig1}
\end{figure*}

Lateral Spine Images (LSIs) are a cornerstone in musculoskeletal diagnostics, providing critical insights for both disease detection and treatment planning. Key applications include spinal alignment assessment to diagnose abnormal curvature conditions such as kyphosis~\cite{kyphosis} and lordosis~\cite{lordosis}, Vertebral Fractures Assessment (VFA)~\cite{vfa}, Bone Mineral Density (BMD) calculation for osteoporosis analysis~\cite{osteoporosis}, Abdominal Aortic Calcification (AAC) detection~\cite{aac1, aac2}, and visceral fat estimation~\cite{maqsood2025pixels}. These LSIs can be acquired through various imaging modalities, including Computed Tomography (CT), Digital X-Ray Imaging (DXI), Magnetic Resonance Imaging (MRI) or Dual-Energy X-ray Absorptiometry (DXA) (refer to Figure~\ref{fig1} for visual comparison). Among these, DXA scans are the fastest, most cost-effective, and low-radiation option, making it the preferred modality for vertebral fracture assessment in routine osteoporosis screening. Traditionally, DXA scans are widely used for BMD measurement~\cite{osteoporosis}. However, there has been growing research interest in leveraging LSI spine DXA scans for additional diagnostic applications. Recent studies have demonstrated their utility in detecting AAC, a stable marker for the development of cardiovascular diseases~\cite{aac1, aac2, showattend, afsah, naeha, zilyas_miccai2024}, as well as in estimating visceral adipose tissue, an important indicator for metabolic health~\cite{maqsood2025pixels}. This trend highlights a growing interest in re-purposing LSI DXA scans beyond bone health toward comprehensive, multi-disease risk profiling.

\begin{figure}
\centerline{\includegraphics[width=0.95\columnwidth]{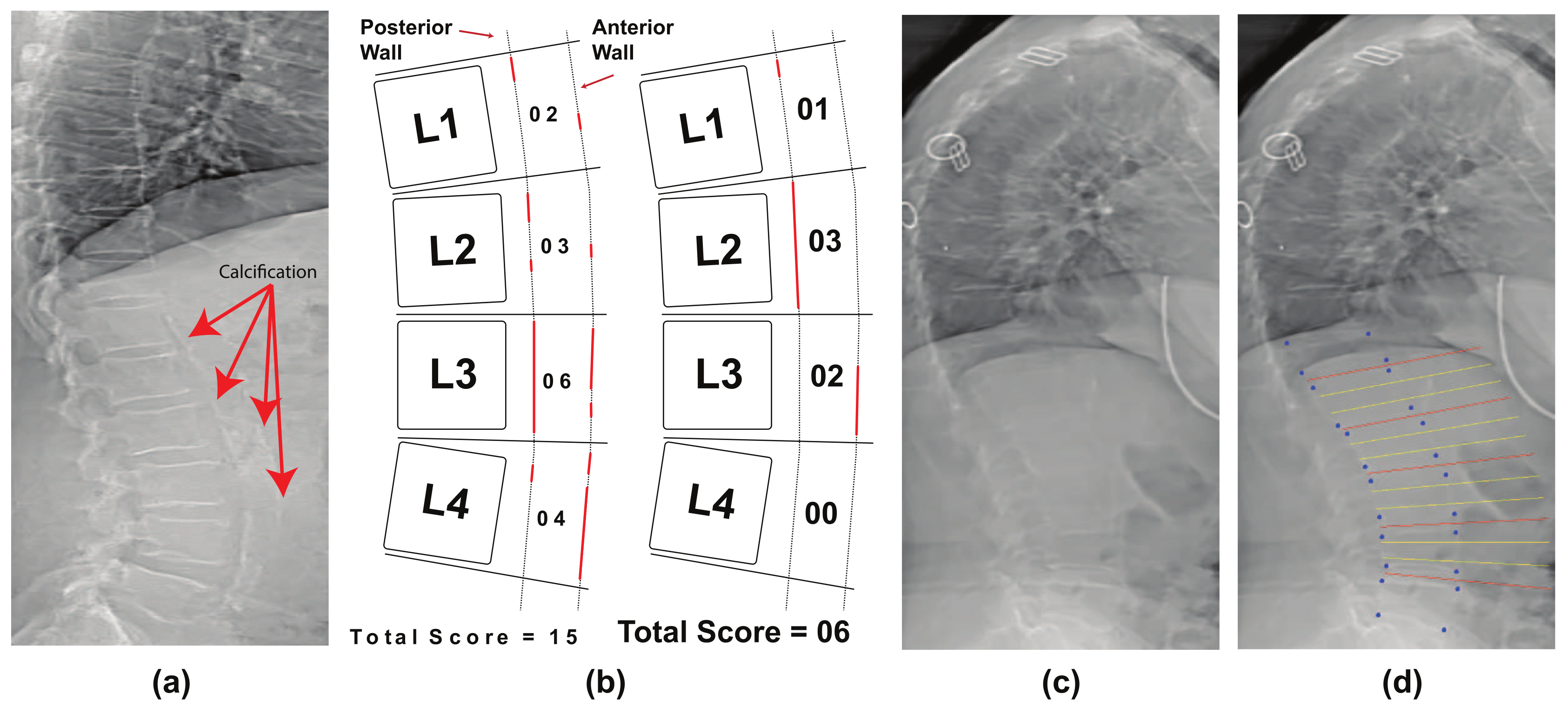}}
\caption{(a) A DXA LSI example with red arrows marking the location of AAC. (b) An illustration of Kauppila's AAC-24 scoring method~\cite{kauppila}. (c) A DXA image example showing unclear vertebral boundaries, with (d) indicating the intended placement of IVGs needed for AAC-24 scoring.}
\vspace{-2mm}
\label{fig2}
\end{figure}

Beyond detection, the quantification of AAC on LSIs typically employs the most widely adopted Kauppila's AAC-24 point scoring method~\cite{kauppila} (refer to Figure~\ref{fig2} for more details). A key component of this method involves identifying the inter-vertebral boundaries between vertebrae (T12, L1, L2, L3, L4 and L5) which require precise vertebral localization. Once these boundaries are determined, AAC is graded on the scale from 0 to 24 (for details, please refer to~\cite{zaid}). This entire AAC-24 scoring process, including vertebral boundary identification (either manually or through digital annotation) and calcification detection, is complex, time-consuming, expensive, and subjective~\cite{afsah, showattend}. This difficulty compounded by factors such as artifacts from kidney stones, bowel gas, and occasionally ambiguous vertebral boundaries. In the context of routine clinical analysis, the scans presented in Figures~\ref{fig1}(c), (d), and (e) illustrate cases where unclear vertebral boundaries complicate the accurate localization of vertebral landmarks. Overcoming these limitations demands extra time and effort for image processing to improve quality and conduct analysis.\par

These challenges have motivated significant research toward automated interpretation of DXA LSIs, focusing on two core tasks: Vertebral Landmark Localization (VLL)~\cite{k_elmasri, t_cootes, wu, sun, yang, payer, zixun, zaid}, and AAC quantification~\cite{k_elmasri, t_cootes, dxapilot, showattend, afsah, naeha, zilyas_miccai2024}. Despite these advancements, both procedures still require significant improvement, especially VLL which remains relatively underexplored and forms the primary focus of this work. Early attempts by Elmasri et al.~\cite{k_elmasri} and Chaplin et al.~\cite{t_cootes} utilized traditional active appearance and shape-based models for vertebral localization but achieved limited success due to small datasets and reliance on handcrafted features. Among few existing studies, Ilyas et al.~\cite{zaid} represent the only study to apply deep learning for VLL in DXA LSIs. Their proposed \lq{GuideNet}\rq, a Convolutional Neural Network (CNN) based model trained on 197 DXA images from a Hologic machine, achieved favourable results but struggled to generalize across scanners from different manufacturers and in cases with complex anatomical structures. This underscores a need for further research into context-aware models that can handle scanner heterogeneity, image noise, and anatomical variability. \par

To accurately categorize or localize components within an image that have a complicated background, low contrast, low signal-to-noise ratio, and artifacts, it is essential to consider both the overall context and the local fine-grained information. For example, without the information of curvature, location, and shape of the spine, vertebral boundaries can be miscategorized as an artifact, or an artifact can be misclassified as AAC. Similarly, without fine-grained textural information, it is difficult to accurately localize the corners of the vertebrae or properly quantify the small amounts of AAC. This can be understood in terms of frequencies embedded in medical images. Low Frequencies (LF) are associated with information such as the shape and curvature of the spine, and High Frequencies (HF) are associated with fine-grained textural information. Both categories of frequency play an important role in informing the decision-making process in image analysis. To the best of our knowledge, the joint utilization of LF and HF components has not been comprehensively investigated for lateral view DXA image analysis, particularly in the context of VLL. Building upon this frequency-based perspective, recent work~\cite{zilyas_miccai2024} introduced the Dual Resolution Self-Attention (DRSA) mechanism for precise AAC quantification. However, directly applying DRSA to VLL tasks fails to effectively capture critical cross-domain contextual information during the feature fusion process at the decoder stage. 

To bridge this gap, we introduce a novel Multi-Context Feature Fusion Block (MCFB), a novel architecture that fuses dual-resolution attention with self- and cross-attention mechanisms at the decoder stage. The MCFB captures spatially enriched contextual features while preserving fine-grained detail, integrating both LF and HF domains. Furthermore, a Channel-wise Self-Attention (CSA)~\cite{restormer, csa2} is employed to model inter-channel dependencies. By operating across multiple hierarchical feature levels, the proposed MCFB effectively integrates LF and HF representations, enabling a more holistic interpretation of the input image. This results in substantial improvements in landmark localization accuracy for DXA LSI scans.
\par

To further demonstrate the clinical utility of accurate VLL in lateral spine DXA images, we propose an algorithm that uses VLL to estimate the Region of Interest (ROI) surrounding the spine, corresponding to the potential location of the abdominal aorta. In the cases where the ROI is partially obscured by the image edge or blackout regions, commonly due to radiation reduction technology in certain DXA scans, our proposed algorithm automatically detects abdominal aorta cropping and quantifies both its location and extent as a percentage. The comparative analysis of our algorithm with expert-annotated images demonstrates the algorithm's accuracy in detecting cropping, while also demonstrating its efficiency in processing large image volumes with minimal human intervention, thereby saving considerable time and computational resources.

We also demonstrate that accurate vertebral landmarks can improve Inter-Vertebral Guide (IVG) generation, which in turn improves the consistency of AAC-24 scoring on DXA LSIs. A preliminary proof-of-concept study yielded encouraging results; however, further validation and clinical trials are necessary. Major contributions in this work are:
\begin{itemize}
    \item A Multi-Context Feature Fusion Block (MCFB) for effective feature fusion at the decoder end.  It utilizes dual resolution attention in both self- and cross-directions with DRSA and DRCA blocks. Additionally, it applies Channel-wise Self-Attention (CSA) to capture interactions between the elements of the combined feature maps.
    \item A hybrid CNN-Transformer architecture `VerteNet', using a pre-trained CNN backbone as the encoder and an MCFB-equipped decoder, achieving state-of-the-art results in Vertebral Landmarks Localization (VLL) for DXA LSI scans.
    \item An abdominal aorta crop detection algorithm leveraging VerteNet’s precise VLL to enable automated detection of abdominal aorta cropping in large datasets. 
    \item A proof of concept showing that IVGs generated by VerteNet could enhance inter-reader agreement among novice readers using the AAC-24 scoring method for detailed AAC evaluation.
\end{itemize}

\section{Related Work}
Numerous studies have attempted to accurately localize vertebral landmarks in spinal imaging; however, research specifically targeting DXA-based LSIs remains scarce. Earlier approaches predominantly relied on hand-crafted features and traditional image processing techniques, limiting their robustness and generalization across diverse anatomical variations and imaging conditions. Elmasri et al.~\cite{k_elmasri} used 56 landmarks to train an active appearance model for localizing the lumbar region and the abdominal aorta in DXA LSIs. Chaplin et al.~\cite{t_cootes} later trained a statistical model on Sagittal CT images using 30 landmarks to locate the lumbar vertebrae, which was then applied to DXA LSIs. Both methods were constrained by the low contrast and resolution of DXA LSIs, leading to poor landmark localization and limited generalization. 

Recently, deep learning-based methods have emerged as the preferred approach, offering significantly improved accuracy and robustness compared to traditional hand-crafted feature techniques. Sun et al.~\cite{sun} developed a multitask framework for VLL in Anterior-Posterior (AP) view X-ray images, while Wu et al.~\cite{wu} proposed BoostNet, which achieved robust VLL by eliminating harmful outliers in the feature space. Yang et al. \cite{yang} developed a VLL framework based on heatmaps and incorporated a Markov Random Field model for refining landmarks. Payer et al.~\cite{payer} proposed a CNN-based model that learns the spatial configuration of landmarks, improving robustness by eliminating the need for heatmap-based localization. Yi et al.~\cite{yi} further advanced this by generating three outputs for landmark localization: center offset, corner offset, and heatmap-based center localization. Guo et al.~\cite{guo} introduced a vertebra segmentation task using a key-point-based transformer architecture to capture the relationships between the global spine structure and local vertebrae. Most recently, Huang et al. \cite{zixun} presented the NFDP model, a generative distribution prior-based model that significantly improved results in AP view X-ray images. Recent studies have also explored anatomical-prior and transformer-based strategies for vertebral landmark localization and spinal deformity assessment in X-ray images, further highlighting the importance of incorporating global spinal context and anatomical consistency into landmark localization frameworks~\cite{batool2025transformer, yang2025anatomical}.

However, the majority of existing research has focused on X-ray images. Aside from the studies by Elmasri et al.~\cite{k_elmasri} and Chaplin et al.~\cite{t_cootes}, the only deep learning-based framework specifically developed \lq{GuideNet}\rq for LSIs was proposed by Ilyas et al.~\cite{zaid}. Their model localized the vertebrae (T12 to L5) to generate IVGs essentially required for AAC-24 scoring but their method was constrained by a limited dataset of 197 DXA scans acquired from a single DXA scanner. Architecturally, the model relies on standard convolutional encoders and simple skip connections without any explicit multi-context or cross-scale features. This limits its ability to integrate global curvature and local edge details, making it vulnerable to low-contrast, noise, and/or artifact-rich DXA images. Additionally, the model does not impose structural constraints to maintain inter-vertebral continuity or anatomical consistency.

Building on recent advances in attention-based architectures, hybrid CNN–Transformer designs have become increasingly common in medical image analysis because they combine the local feature extraction strength of CNNs with the long-range contextual modelling capability of transformers~\cite{khan2025visiontransformers}. Transformer-based models have shown strong potential for capturing complex spatial dependencies in medical imaging. While several studies have applied such models for vertebral landmark localization in anterior–posterior (AP) X-ray views~\cite{guo, moxin}, their application to lateral DXA images remains largely unexplored.

To overcome the above-mentioned limitations, we propose VerteNet, a hybrid CNN–Transformer that introduces multi-context feature fusion block that effectively integrates information across multiple feature scales. Additionally, by leveraging the added two attention mechanisms, VerteNet enhances both local detail and global anatomical context, ensuring spatial continuity and anatomical coherence across vertebrae (T12 to L5). To the best of our knowledge, VerteNet is among the first transformer-based frameworks specifically developed and evaluated for vertebral landmark localization in lateral DXA spine images.

\section{Methodology}
\subsection{Overall Framework}
\begin{figure*}
\centerline{\includegraphics[width=0.88\textwidth]{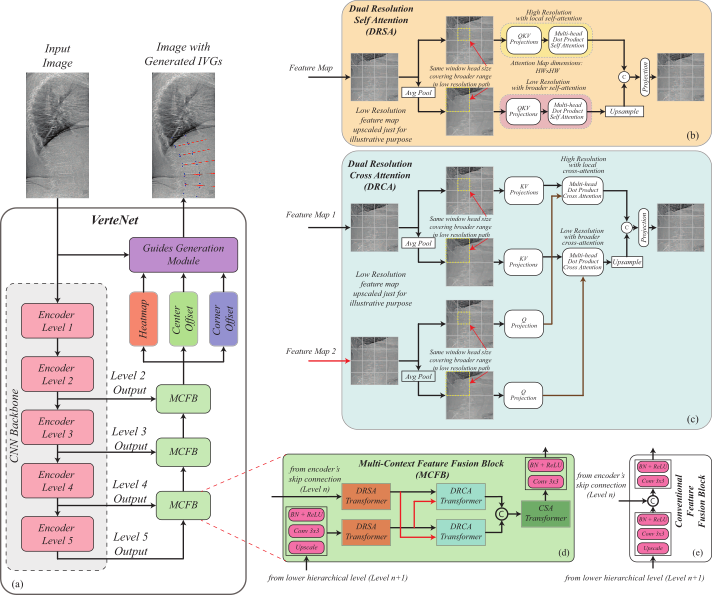}}
\caption{(a) Proposed Framework VerteNet (b) Dual Resolution Self Attention (DRSA) (c) Dual Resolution Cross Attention (DRCA) - takes two feature maps as input and generates Query from one feature map and Keys and Values from the other feature map. (d) Multi-Context Feature Fusion Block (MCFB) that employs DRSA and DRCA to calculate self-attention within, and cross attention among features from skip connection of layer \textit{n} and upscaled feature map from layer \textit{n}+1. Next, it utilizes a channel-wise self-attention transformer block to calculate inter-correlation among channels. (e) Conventional feature fusion approach that simply concatenates features along channels, and then performs convolution operations with activation functions (lack both inter and intra-correlation information among elements of the feature maps).}
\label{fig3}
\end{figure*}

We propose VerteNet, a hybrid CNN-Transformer architecture for vertebral landmark localization in lateral spine DXA scans. The proposed model comprises four key components : (a) an encoder that extracts multi-scale spatial representations using a pre-trained convolutional backbone, (b) a decoder equipped with a novel Multi-Context Feature Fusion Block (MCFB) that integrates dual-resolution attention mechanisms, (c) output regression heads that generate vertebral heatmaps, center offsets, and corner offsets for precise landmark prediction, and (d) an auxiliary aorta cropping detection module that leverages the localized landmarks to automatically identify incomplete abdominal regions in DXA scans. The detailed architecture is shown in Fig.~\ref{fig3}(a). 

Unlike conventional convolutional frameworks that rely solely on local receptive fields, the proposed VerteNet combines CNN for low-level feature extraction with Transformer-based attention blocks for capturing long-range dependencies. The encoder employs a pretrained convolutional backbone (i.e. EfficientNetV2S~\cite{tan2021efficientnetv2}) that progressively extracts multi-scale features through convolutional and non-linear activation operations. Down-sampling layers reduce the spatial resolution while increasing the feature-map depth, resulting in a compact representation of the input image while preserving essential features. Feature maps from multiple encoder stages are fed to decoder via skip-connections to preserve high-frequency spatial information. The decoder then performs progressive up-sampling and feature fusion to reconstruct dense localization maps while integrating the encoder’s contextual cues.

The novelty in VerteNet resides in its feature-fusion strategy (i.e. MCFB) implemented within the decoder. Unlike the conventional simple fusion approach shown in Fig.~\ref{fig3}(e), where feature maps are simply concatenated along the channel dimension and adjusted through convolution, MCFB addresses the limitations of simple fusion. Simple fusion lacks critical information about global interactions among feature map elements in both spatial and channel domains, doesn’t consider the importance of elements in one feature map relative to the other, and overlooks the different frequency information inherent in an image.

MCFB overcomes these limitations by using DRSA and DRCA at different feature map resolutions, along with the concept of CSA~\cite{restormer, csa2}, as shown in Fig.~\ref{fig3}(d). These modules, which will be explained in the following sections, have proven effective and significantly improved VLL results in DXA LSIs.

For the regression heads, VerteNet also follows the same approach as GuideNet~\cite{zaid} for heatmap, center offset, and corner offset calculations. The heatmap module provides the approximate location of each vertebra's center, the center offset module refines this location, and the corner offset module calculates the coordinates of the vertebrae corners. For more details on these modules, please refer to GuideNet~\cite{zaid}. 
The following subsections will first explain DRSA, then the working principle of DRCA, and finally the details of MCFB.       

\subsection{Dual Resolution Self Attention (DRSA)}
Self-Attention (SA) has proven effective in capturing long-range dependencies and inter-correlations among elements within a single feature map~\cite{attention, vit, hilosa}. However, it still misses important information due to some limitations of attention calculation. Usually, SA is calculated using the concept of patches to control the quadratic complexity of attention calculation. This limits the receptive field of SA limiting the context of these operations. In natural images, including medical images, a broad range of frequencies exists holding essential details about the objects in the image. HFs relate to fine-grained details like texture, while LFs are related to global structures, such as the size, shape, and curvature of the object. Capturing all these details is essential, but simple patch-wise attention mechanisms with limited context lose global context ~\cite{swin, dattention}. On the other hand, increasing the patch size to extend the context of SA increases the computational cost and parameter count quadratically. DRSA is designed to address these challenges. The first objective is to capture the diverse frequency information inherent in DXA images, and the second objective is to expand the spatial context of the windowed SA. DRSA splits the conventional Multi-Head Self-Attention (MHSA) into two sections. One section uses local windowed-SA on the input feature map, on its actual size, to encode the HF interactions, and the other section first downsamples the input feature map using average pooling operation which is a low-pass filter operation in nature, and then applies windowed SA on it. This section captures the information about the LF interaction. We use the same-sized windows in both DRSA sections which increases the spatial context of windowed SA in LF section and also introduces an overlap when the encoded feature maps are combined eventually. The DRSA block has been illustrated in Fig.~\ref{fig3}(b). Given an input feature map \textit{X} $\in$ $\mathbb{R}^{H \times W \times C}$, where \textit{H}, \textit{W}, and \textit{C} are the height, width, and channels of the input feature map, respectively, the DRSA down-samples it using avg. pooling to generate a low-resolution feature map \textit{X'} $\in$ $\mathbb{R}^{H/r \times W/r \times C}$, where \textit{r} is the reduction factor. Separate Query (Q), Key (K), and Value (V) embeddings are generated for \textit{X} and \textit{X'} using linear layers. Considering a single head, the SA for \textit{X} and \textit{X'}, abbreviated as $\textit{SA}_{H}$ and $\textit{SA}_{L}$ respectively are calculated as follows:
\begin{equation}
\textit{SA}_{H} = \text{softmax}\left(\frac{Q_HK_H^T}{\beta_{1}}\right) V_H
\end{equation}
\begin{equation}
\textit{SA}_{L} = \text{softmax}\left(\frac{Q_LK_L^T}{\beta_{2}}\right) V_L
\end{equation}

where $Q_H$, $K_H$, $V_H$, $Q_L$, $K_L$, and $V_L$ are the Query, Key, and Value embeddings for \textit{X} and \textit{X'} feature maps, with respective self-attentions $\textit{SA}_{H}$ and $\textit{SA}_{L}$. $\beta_{1}$ and $\beta_{2}$ are learnable parameters that control the SA. 

\begin{figure*}[!t]
\centerline{\includegraphics[width=0.75\textwidth]{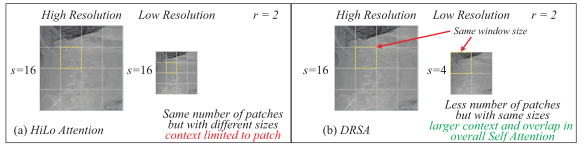}}
\caption{Comparison of (a) HiLo Attention~\cite{hilosa} and (b) DRSA. HiLo uses the same window count \textit{s} across resolutions, limiting context, whereas DRSA uses the same window size to expand low-resolution context and introduce overlap. Here, \textit{r} is the reduction factor and \textit{s} is the window count.}
\vspace{-2mm}
\label{fig4}
\end{figure*}

\begin{equation}
\textit{SA}_{Total} = \text{Concat}(\textit{SA}_{H}, \text{Upsample}(\textit{SA}_{L}))W_p
\end{equation}

where \textit{SA$_{Total}$} is the overall self-attention, calculated by first up-sampling the $\textit{SA}_{L}$ using a transposed convolution layer (learnable), then concatenating it with $\textit{SA}_{H}$, and finally passing it through a linear layer $W_p$.

The DRSA is inspired by HiLo SA~\cite{hilosa}, but it uses a different approach to cover the limitations of HiLo SA, which was mainly designed for fast transformer models. The architectural design of DRSA covers the limitation of HiLo SA, i.e. Firstly, HiLo SA uses the same window size in both Hi and Lo attention sections, which restricts spatial context. Secondly, the HiLo SA avoids overlap of patches while ``paying attention''. The working principle comparison of DRSA and the HiLo SA is shown in Fig.~\ref{fig4}.

\subsection{Dual Resolution Cross Attention (DRCA)}
Self-attention captures long-range dependencies and inter-correlations among elements within a single feature map. However, when combining two feature maps, the concept of SA can be extended in a cross direction i.e. \textit{Q} embeddings are generated from one feature map, and \textit{K} and \textit{V} embeddings are generated from the other feature map. This concept of Cross-Attention (CA) plays a critical role in managing inter-correlations and inter-dependencies between elements from different feature maps~\cite{attention, bert}. Instead of using a naive CA approach, we propose DRCA which utilizes the similar concept of dual resolution attention in the CA domain. The DRCA block has been illustrated in Fig.~\ref{fig3}(c). Given two input feature maps \textit{X} and \textit{Y}, both of the shape $\mathbb{R}^{H \times W \times C}$, DRCA splits the conventional MHSA into two sections. In one section, the DRCA down-samples the feature maps \textit{X} and feature maps \textit{Y} using avg. pooling to generate a low-resolution feature map \textit{X'} and \textit{Y'} $\in$ $\mathbb{R}^{H/r \times W/r \times C}$, where \textit{r} is the reduction factor. Separate Query embeddings, $\textit{Q}^{X}_{H}$ and $\textit{Q}^{X'}_{L}$, are generated for \textit{X} and \textit{X'}. Now different from DRSA, Key and Value embeddings are generated from feature map \textit{Y} i.e. $\textit{K}^{Y}_{H}$, $\textit{K}^{Y'}_{L}$, $\textit{V}^{Y}_{H}$, $\textit{V}^{Y'}_{L}$. Considering a single head, the CA abbreviated as $\textit{CA}_{H}$ and $\textit{CA}_{L}$ respectively are calculated as follows:

\begin{equation}
\textit{CA}_{H} = \text{softmax}\left(\frac{Q^{X}_{H}K^{Y^{T}}_{H}}{\beta_{1}}\right)V^{Y}_{H}
\end{equation}
\begin{equation}
\textit{CA}_{L} = \text{softmax}\left(\frac{Q^{X'}_{L}K^{Y'^{T}}_{L}}{\beta_{2}}\right)V^{Y'}_{L}
\end{equation}

where $\textit{Q}^{X}_{H}$, and $\textit{Q}^{X'}_{L}$ are the Query embeddings generated from feature map \textit{X} and its scaled down version \textit{X'} respectively. $\textit{K}^{Y}_{H}$, $\textit{K}^{Y'}_{L}$, $\textit{V}^{Y}_{H}$, $\textit{V}^{Y'}_{L}$ are the Key and Value embeddings generated from the feature map \textit{Y} and its scaled down version \textit{Y'}. $\beta_{1}$ and $\beta_{2}$ are learnable parameters that control the CA. After calculating $\textit{CA}_{H}$ and $\textit{CA}_{L}$, the overall cross attention \textit{$CA_{Total}$} is calculated by first up-sampling the $\textit{CA}_{L}$ using a transposed convolution layer (learnable), then concatenating it with $\textit{CA}_{H}$, and finally passing it through a linear layer $W_p$.   
\begin{equation}
\textit{CA}_{Total} = \text{Concat}(\textit{CA}_{H}, \text{Upsample}(\textit{CA}_{L}))W_p
\end{equation}

\begin{figure*}
\centerline{\includegraphics[width=0.94\textwidth]{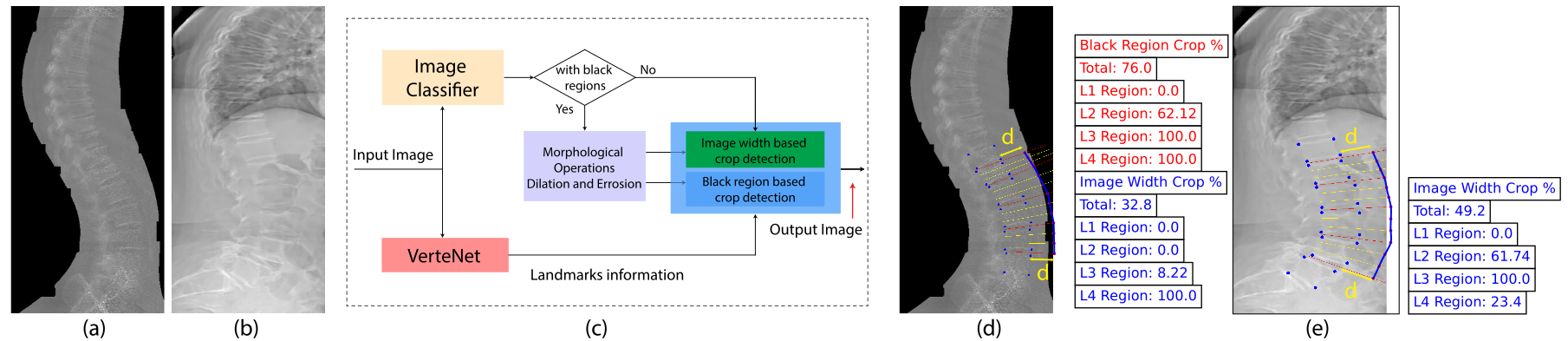}}
\caption{(a) and (b) describe the types of input images that our algorithm can process, originating from two different machines: one with black regions (GE machine) and one without (Hologic machines). (c) illustrates the block-level structure of the abdominal aorta crop detection algorithm. The image classifier categorizes the input images into two groups: those with black regions and those without (d). (e) shows the corresponding outputs for image types (a) and (b), indicating both the location and the percentage of the abdominal aorta crop detected.}
\vspace{-2mm}
\label{fig_algo}
\end{figure*}

\subsection{Multi-Context Feature Fusion Block (MCFB)}
For effective utilization of DRSA and DRCA modules when combining the feature maps from skip connections and the lower layers of the decoder module, an efficient approach is required that incorporates multi-context information along with rich inter and intra-correlation information. So, we propose MCFB, which effectively utilizes DRSA and DRCA in addition to CSA for this purpose. It is shown in Fig.~\ref{fig3}(a) and explained in Fig.~\ref{fig3}(d). It takes feature maps from the lower layer \textit{n}+1 of the decoder part and the skip connection from the encoder layer \textit{n}. First, it upsamples the feature map from \textit{n}+1 using interpolation, and performs a convolutional operation followed by an activation function. MCFB passes both the feature maps from independent DRSA-based transformer blocks. Next, MCFB passes the feature maps through DRCA-based independent transformer blocks. After adding both self and cross-dual resolution attention, MCFB concatenates the feature maps along the channel direction. To find inter-correlation among the channels of the concatenated feature maps, MCFB passes the resultant feature map through a transformer block with transposed self-attention/ CSA, adding channel-wise self-attention. Given two feature maps $\text{X}_{S}$ and $\text{X}_{D}$ $\in$ $\mathbb{R}^{H \times W \times C}$, where $\text{X}_{S}$ is the feature map from the skip connection and $\text{X}_{D}$ is the feature map from the decoder layer after upsampling, convolution, and activation function, the operation of MCFB is as follows:
\begin{equation}
\begin{split}
\textit{F}_{S} = \textit{GDFN}(\textit{DRSA}(\textit{X}_{S})W_s + \textit{X}_{S})W_p + (\textit{DRSA}(\textit{X}_{S})W_s + \\
\textit{X}_{S})    
\end{split}
\end{equation}
\begin{equation}
\begin{split}
\textit{F}_{D} = \textit{GDFN}(\textit{DRSA}(\textit{X}_{D})W_d + \textit{X}_{D})W_q + (\textit{DRSA}(\textit{X}_{D})W_d + \\
\textit{X}_{D})    
\end{split}
\end{equation}
\begin{equation}
\begin{split}
\textit{F}_{SD} = \textit{GDFN}(\textit{DRCA}(\textit{F}_{S},\textit{F}_{D})W_a + \textit{F}_{S})W_p' + \\
(\textit{DRCA}(\textit{F}_{S},\textit{F}_{D})W_a + \textit{F}_{S})    
\end{split}
\end{equation}
\begin{equation}
\begin{split}
\textit{F}_{DS} = \textit{GDFN}(\textit{DRCA}(\textit{F}_{D},\textit{F}_{S})W_b + \textit{X}_{D})W_q' + \\
(\textit{DRCA}(\textit{F}_{D},\textit{F}_{S})W_b + \textit{X}_{D})
\end{split}
\end{equation}
\begin{equation}
\textit{F}_{C} = \textit{Concat}(\textit{F}_{SD}, \textit{F}_{DS})W_r
\end{equation}

where $\textit{F}_{S}$ and $\textit{F}_{D}$ are the feature maps $\text{X}_{S}$ and $\text{X}_{D}$ after passing through DRSA based transformer blocks. $\textit{F}_{SD}$ is the feature map $\textit{F}_{S}$ after incorporating dual resolution cross attention with respect to $\textit{F}_{D}$ through transformer block with DRCA. Similarly, $\textit{F}_{DS}$ is the feature map $\textit{F}_{D}$ after incorporating dual resolution cross attention with respect to $\textit{F}_{S}$, through a separate transformer block with DRCA. $\textit{F}_{C}$ is the resultant feature map after concatenation of $\textit{F}_{DS}$ and $\textit{F}_{SD}$. The resultant feature map $\textit{F}_{C}$ has multi-context spatial information added to it using DRSA and DRCA operations, however, for calculating inter-correlation of elements among channels direction, a modified version of SA i.e. CSA is used. Given the generated \textit{Q}, \textit{K}, \textit{V} $\in$ $\mathbb{R}^{2C \times HW}$ embeddings generated from the feature map $\textit{F}_{C}$ $\in$ $\mathbb{R}^{H \times W \times 2C}$, the CSA is calculated using the following equation:

\begin{equation}
   \textit{CSA} = V^{T} \cdot \text{softmax}\left(\frac{KQ^{T}}{\beta}\right)
\end{equation}

The overall equation for the transformer block employing CSA is as follows:

\begin{equation}
\textit{F}_{O} = \textit{GDFN}(\textit{CSA}(\textit{F}_{C})W_t + \textit{F}_{C})W_o + (\textit{CSA}(\textit{F}_{C})W_t + \textit{F}_{C})
\end{equation}
\begin{equation}
    F_{O'} = \textit{ReLU}(\textit{BN}(\textit{F}_{O}W_c))
\end{equation}

Lastly, the MCFB block passes the resultant $\textit{F}_{O}$ through the convolutional layer $W_c$, batch normalization layer, and ReLU activation layer to generate final output $\textit{F}_{O'}$. In all the above equations $W_s$, $W_p$, $W_d$, $W_q$, $W_a$, $W_p'$, $W_b$, $W_q'$, $W_r$, $W_o$, and $W_t$ are the linear layers with learnable parameters. $\textit{GDFN}$, used in DRSA, DRCA and CSA, is Gated-Dconv Feed-Forward Network~\cite{restormer} that regulates the flow of information to concentrate on details that complement the others.

\subsection{Abdominal Aorta Crop Detection Algorithm}
Based on the outputs of `VerteNet', we propose an algorithm that accurately determines whether the input image contains a cropped abdominal aorta. The algorithm, depicted in Fig.~\ref{fig_algo}(c), facilitates the efficient analysis of large cohorts of VFA images acquired from various DXA machines, thereby significantly reducing the workload of image readers. It is compatible with images from both GE machines, shown in Fig.~\ref{fig_algo}(a), which exhibit black regions due to radiation reduction technology, and conventional Hologic machines, illustrated in Fig.~\ref{fig_algo}(b). In addition to detecting abdominal aorta cropping, the algorithm quantifies the percentage and identifies the specific location of cropping. The detailed steps of the algorithm for detecting potential abdominal aorta cropping are provided below:

\begin{itemize}
    \item \textbf{Image Categorization and Pre-processing:} The input image is processed by a CNN-based classifier to determine if it contains black regions. If black regions are detected, a pre-processing step applies dilation and erosion to smooth the black regions.
    \item \textbf{Determine Line Equations:} The equations for lines connecting anterior and posterior landmarks of inter- (red) and intra- (yellow) vertebral guides are computed (see images Fig.~\ref{fig_algo}(d) and Fig.~\ref{fig_algo}(e)).
    
    \item \textbf{Find Points at Distance $d$:} Using these line equations, the coordinates and pixel values of points located at a predefined distance \(d\) from all anterior landmarks (red dots in images Fig.~\ref{fig_algo}(d) and Fig.~\ref{fig_algo}(e)) are determined. The distance \(d\) is defined as a function of the mean vertebral width, i.e., \(d = \textit{Factor} \times (1+\textit{Mean Vertebral Width})\), with \textit{Factor} being a predefined constant.
    
    \item \textbf{Fit a Cubic Spline:} A cubic spline is fitted to the points at a distance $d$ from the anterior landmarks to derive the spline equation. 

    \item \textbf{Calculate Pixel Values:} Using the cubic spline equation, the pixel values for $x$ number of evenly spaced points (e.g., 500 points) are calculated, represented as a blue line in images Fig.~\ref{fig_algo}(d) and Fig.~\ref{fig_algo}(e).
    
    \item \textbf{Check for Cropping in Image Fig.~\ref{fig_algo}(e):} For cases like image (e), if the coordinates of any of the $x$ points exceed the image width, this indicates potential abdominal aorta cropping, as shown in the image Fig.~\ref{fig_algo}(e).
    
    \item \textbf{Check for Cropping in Image Fig.~\ref{fig_algo}(d):} For cases like image Fig.~\ref{fig_algo}(d), if the image contains black regions, check two conditions:  
    \begin{itemize}
        \item Whether the pixel values of any of the $x$ points fall within these black regions.  
        \item Whether the coordinate values of any of the $x$ points exceed the image width.
    \end{itemize}
    In either case, this suggests a possibility of abdominal aorta cropping. Finally, the percentage and location of the cropping is calculated.
    
\end{itemize}

\section{Experimental Setup}
\subsection{Dataset Used}
De-identified images were sourced from vertebral fracture assessment studies, with ethics approvals obtained from relevant boards, including HealthPartners, Edith Cowan University, the University of Manitoba, and the Manitoba Health Information Privacy Committee (details on the first page). We selected and annotated 620 scans from three different machines, i.e. 200 scans from the iDXA GE machine (100 SE and 100 DE), 100 SE scans from Hologic 4500A, and 320 SE scans from the Hologic Horizon machine.

Annotation was performed using an offline version of the makeSense\footnote{https://www.makesense.ai/} tool, where the four corner points of each vertebra (T12 to L5) were manually marked to serve as ground-truth vertebral landmarks for training and evaluation.

\subsection{Implementation Details}
The SE and DE image sets from the GE iDXA represent paired scans from the same patients, yet the images were produced using distinct acquisition methodologies, leading to different pixel value distributions across the two sets. Hence, all images were rescaled to a uniform distribution of 1024$\times$512 pixels. The proposed model was implemented in PyTorch and trained on an NVIDIA A6000 GPU. We performed 10-fold cross-validation, ensuring that each fold preserved the same proportional distribution of scans from all three DXA machines. For training the model, batch size of 12 and a learning rate of 0.0001 was used. To mitigate overfitting and improve model generalization, we applied several data augmentations techniques i.e. random cropping, random expanding, brightness, and contrast distortion. We used Focal Loss~\cite{focalloss} for Heatmap output and L1 loss for Center and Corner Offset outputs. For evaluation normalized mean error and normalized median error were employed to assess landmark localization accuracy, while binary classification accuracy between `cropped' and `not cropped' categories was used to evaluate aorta crop detection performance.

\section{Experimental Results and Discussion}
\subsection{Landmark Localization Results}
To the best of our knowledge, the only prior work in VLL domain is GuideNet~\cite{zaid}, which was initially trained on a limited dataset comprising 197 DXA scans acquired from the Hologic machine. For a fair evaluation, we retrained GuideNet on the same dataset used to train VerteNet and conducted a comparative performance analysis. Additionally, we evaluated the performance of our model against HRNet~\cite{hrnet} and NFDP~\cite{zixun}. HRNet, a well-established framework for landmark localization, was originally designed for human pose estimation but has also been widely applied in medical image landmark localization tasks. NFDP, a recent state-of-the-art (SOTA) model, was primarily trained on diverse X-ray images, including AP view X-ray spine images, X-ray cephalograms, and X-ray hand images. NFDP's SOTA performance in these tasks can be attributed to its innovative use of generative distribution priors. To ensure a fair comparison, we trained both HRNet and NFDP on the same dataset used for VerteNet and subsequently compared their performances. As shown in Table~\ref{table 1}, our proposed model outperformed these models.

\begin{table}[H]
\caption{Landmark Localization Results}
\label{table 1}
\begin{center}
\begin{tabular}{lcc}
\hline
\textbf{Method} & \textbf{\begin{tabular}[c]{@{}c@{}}Normalized \\ Mean Error\\ (Pixels)\end{tabular}} & \textbf{\begin{tabular}[c]{@{}c@{}}Normalized \\ Median Error\\ (Pixels)\end{tabular}} \\ \hline
GuideNet~\cite{zaid}        & 7.78                                                                                 & 2.96                                                                                   \\
HRNet~\cite{hrnet}           & 5.67                                                                                 & 2.49                                                                                   \\
NFDP~\cite{zixun}             & 5.02                                                                                 & 2.44                                                                                   \\ \hline
VerteNet (Ours) & 4.92                                                                                 & 2.35                                                                                   \\ \hline
\end{tabular}
\end{center}
\end{table}

\subsection{Abdominal Aorta Crop Detection Results}
\begin{figure*}
\centerline{\includegraphics[width=0.72\textwidth]{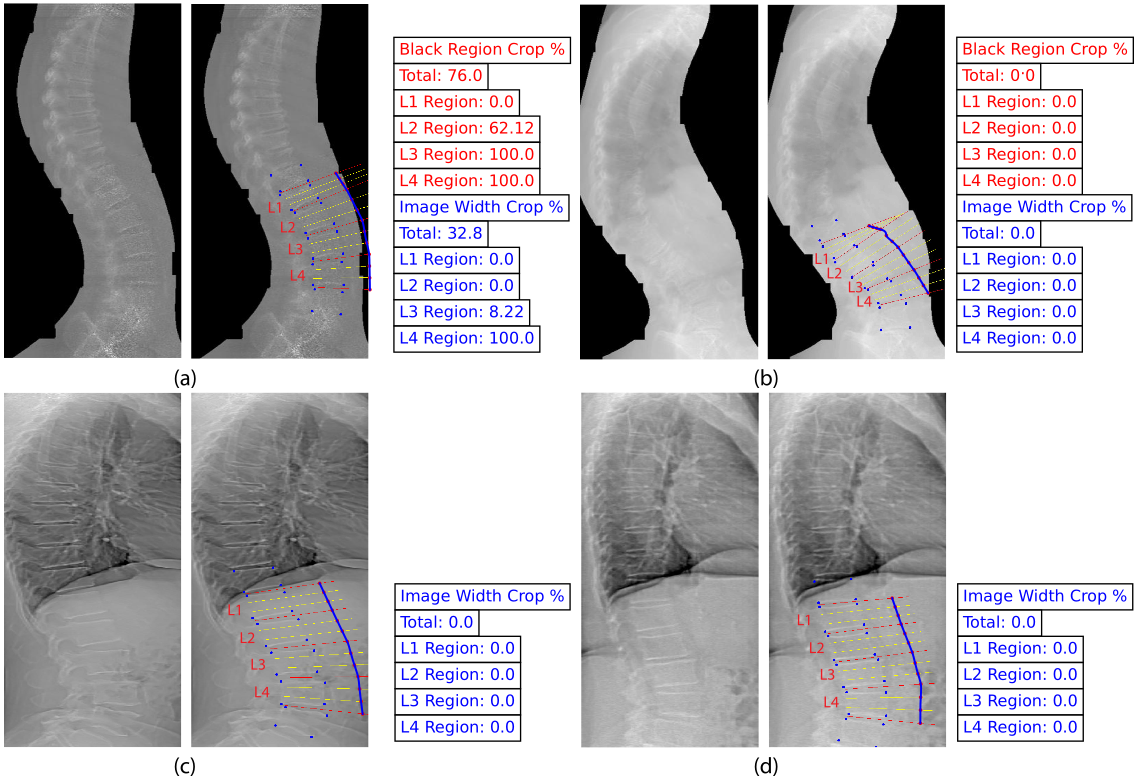}}
\caption{Examples of abdominal aorta crop detection. (a) Crop detected in a GE DE DXA image using black-region and image-width criteria. (b) No crop detected from L1 to L4 in a GE SE DXA image. (c,d) No crop detected in SE Hologic images.}
\vspace{-3mm}
\label{fig_results}
\end{figure*}

\begin{figure*}
\centerline{\includegraphics[width=0.7\textwidth]{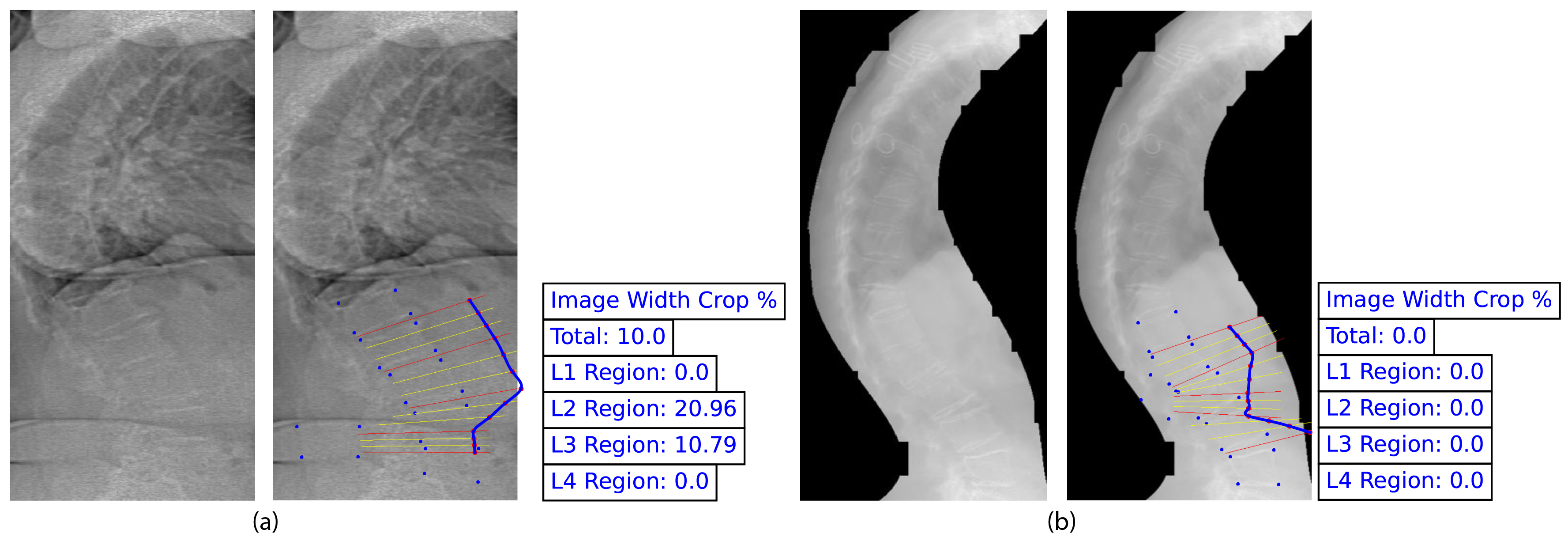}}
\caption{(a) and (b) represent the failure cases of the proposed algorithm, where improper localization of landmarks in images from Hologic and GE machines leads to suboptimal performance, as the algorithm relies on accurate landmark detection.}
\vspace{-3mm}
\label{fig_failure}
\end{figure*}

To comprehensively assess the performance of our algorithm, we conducted two experiments. In the first experiment, we selected 70 labeled images from the Hologic Horizon machine: 35 labeled as having insufficient soft tissue on the LSI (with a cropped aorta) and 35 labeled as having adequate soft tissue for evaluating AAC-24 on the LSI. The labeling was performed by an expert clinician (J.T.S.) with over 15 years of experience in analyzing DXA LSIs. We tested our method with \textit{Factor} values ranging from 0.8 to 1.5 (as shown in Table~\ref{table 3}) and found that a factor value of 1.2 achieved the best agreement with expert labels. In the second experiment, we applied our algorithm to 200 images and, without disclosing the algorithm’s results, asked J.T.S. to independently classify the images as either having a cropped abdominal aorta or not. A comparison of the results revealed strong alignment, confirming the accuracy of the algorithm's predictions at a factor value close to 1.0. The results are presented in Table~\ref{table 3_1}. Fig.~\ref{fig_results} showcase examples of successful cases where our algorithm performed accurately. However, the algorithm depends on accurate landmark placement as shown in Fig.~\ref{fig_failure}.

\begin{table}[H]
\caption{Experiment 1: Comparison of the proposed algorithm's performance with human-labeled abdominal aorta crop detection on 70 Hologic DXA images. True positives (TP) correctly identify cropped aortas, true negatives (TN) identify uncropped aortas, false positives (FP) misclassify uncropped as cropped, and false negatives (FN) misclassify cropped as uncropped.}
\label{table 3}
\begin{center}
\begin{tabular}{cccccc}
\hline
\textbf{Factor} & \textbf{FP} & \textbf{FN} & \textbf{TP} & \textbf{TN} & \textbf{Accuracy ($\%$)} \\ \hline
0.8 & 8 & 0 & 27 & 35 & 88.57 \\
0.9 & 3 & 0 & 32 & 35 & 95.71 \\
1.0 & 2 & 0 & 33 & 35 & 97.14 \\    
1.1 & 1 & 0 & 34 & 35 & 98.57 \\ \hline
1.2 & 0 & 0 & 35 & 35 & 100 \\ \hline
1.3 & 0 & 1 & 35 & 34 & 98.57 \\ 
1.4 & 0 & 3 & 35 & 32 & 95.74 \\ 
1.5 & 0 & 8 & 35 & 27 & 88.57 \\ \hline
\end{tabular}
\vspace{-3mm}
\end{center}
\end{table}

\begin{table}[]
\caption{Experiment 2: Comparison of the performance of the proposed algorithm with human-labeled abdominal aorta crop detection on a dataset of 200 Hologic DXA images.}
\label{table 3_1}
\begin{center}
\resizebox{\columnwidth}{!}{
\begin{tabular}{ccccc}
\hline
\textbf{Factor} & \textbf{Accuracy (\%)} & \textbf{Sensitivity} & \textbf{Specificity} & \textbf{F1-Score} \\ \hline
0.9  & 86.0  & 0.68 & 0.98  & 0.25 \\ \hline
\textbf{1.0} & \textbf{96.0}  & \textbf{0.93} & \textbf{0.98} & \textbf{0.95}  \\ \hline
1.1 & 92.0 & 1.0 & 0.87 & 0.92 \\ \hline
\end{tabular}}
\end{center}
\end{table}

\subsection{Impact of IVGs on Inter-Reader Agreement in Granular AAC Scoring}
This proof-of-concept study investigated whether IVGs improve the correlation between two readers with limited training (2-day course) in evaluating granular AAC scores from DXA LSIs. Readers typically use manual or imaginary IVGs to divide the abdominal aorta into four regions corresponding to L1–L4 vertebrae. While experienced professionals can position IVGs and categorize regions accurately, less experienced readers may misclassify regions, e.g., assigning calcification to L3 instead of L4, or L2 instead of L1. Such errors are unlikely to affect the overall AAC score (0–24) however, it definitely would affect granular scores. In this study, two authors (A.M. and A.S.) with limited training in Kauppila's AAC-24 scoring evaluated 32 DXA images from a Hologic Horizon machine twice: first without IVGs and later with IVGs. Table~\ref{table 2} and Table~\ref{table 2_1} present experimental results, showing improved inter-reader agreement in correlation and Cohen's weighted kappa values.

\begin{table}[h]
\caption{Correlation Coefficient - Inter-Reader}
\label{table 2}
\begin{center}
\resizebox{\columnwidth}{!}{
\begin{tabular}{cclcl}
\hline
\multicolumn{5}{c}{\textbf{Correlation Coefficient - Inter-Reader}} \\ \hline
\multicolumn{1}{l}{} & \multicolumn{2}{c}{\textbf{without IVGs}} & \multicolumn{2}{c}{\textbf{with IVGs}}  \\ \hline
\textbf{\begin{tabular}[c]{@{}c@{}}Vertebral Region \\ (AAC Score Range: 0-6)\end{tabular}} & \multicolumn{2}{c}{\textbf{Value (95$\%$ CI)}} & \multicolumn{2}{c}{\textbf{Value (95$\%$ CI)}} \\ \hline
L1 & \multicolumn{2}{c}{0.82 (0.64 - 0.91)} & \multicolumn{2}{c}{\textbf{0.94 (0.89 - 0.97)}} \\
L2 & \multicolumn{2}{c}{0.88 (0.75 - 0.94)} & \multicolumn{2}{c}{\textbf{0.90 (0.79 - 0.95)}} \\
L3 & \multicolumn{2}{c}{0.83 (0.66 - 0.92)} & \multicolumn{2}{c}{\textbf{0.84 (0.68 - 0.92)}} \\
L4 & \multicolumn{2}{c}{0.91 (0.81 - 0.96)} & \multicolumn{2}{c}{\textbf{0.94 (0.89 - 0.97)}} \\ \hline
\end{tabular}}
\end{center}
\end{table}

\begin{table}[h]
\caption{Cohen's weighted kappa - Inter-Reader}
\label{table 2_1}
\renewcommand{\arraystretch}{1.1}
\begin{center}
\resizebox{\columnwidth}{!}{
\begin{tabular}{cclcl}
\hline
\multicolumn{5}{c}{\textbf{Cohen's weighted kappa - Inter-Reader}} \\ \hline
\multicolumn{1}{l}{} & \multicolumn{2}{c}{\textbf{without IVGs}} & \multicolumn{2}{c}{\textbf{with IVGs}} \\ \hline
\textbf{\begin{tabular}[c]{@{}c@{}}Vertebral Region \\ (AAC Score Range: 0-6)\end{tabular}} & \multicolumn{2}{c}{\textbf{Value (95$\%$ CI)}} & \multicolumn{2}{c}{\textbf{Value (95$\%$ CI)}}     \\ \hline
L1 & \multicolumn{2}{c}{0.57 (0.24 - 0.91)} & \multicolumn{2}{c}{\textbf{0.82 (0.54 - 1.09)}} \\
L2 & \multicolumn{2}{c}{0.54 (0.29 - 0.79)} & \multicolumn{2}{c}{\textbf{0.58 (0.33 - 0.84)}} \\
L3 & \multicolumn{2}{c}{0.57 (0.37 - 0.76)} & \multicolumn{2}{c}{\textbf{0.61 (0.41 - 0.80)}} \\
L4 & \multicolumn{2}{c}{0.66 (0.49 - 0.82)} & \multicolumn{2}{c}{\textbf{0.78 (0.63 - 0.92)}} \\ \hline
\end{tabular}}
\vspace{-3mm}
\end{center}
\end{table}

\subsection{Ablation Studies}
We conducted ablation experiments to validate our architecture. First, we analyzed the impact of the reduction factor \textit{r} and patch size \textit{p} in DRSA and DRCA. Table~\ref{table 4} shows various \textit{r} and \textit{p} combinations at hierarchical levels. Using a larger reduction factor (\textit{r} = 4) in shallower layers significantly reduced performance due to excessive down-sampling and loss of information. For patch size, the best results were obtained with \textit{p} = 10. Larger patch sizes increased parameters and risked overfitting due to insufficient training data. Next, we evaluated different CNN backbones. As shown in Table~\ref{table 5}, EfficientNetV2S outperformed other backbones. Finally, we tested different decoder configurations: a basic U-Net design, DRSA-only, DRCA-only, and our proposed MCFB (DRSA, DRCA, and CSA). Table~\ref{table 6} shows that MCFB achieved the best performance.

\subsection{Performance Comparison with SOTA Models}
To evaluate the efficiency and computational cost of VerteNet, we compared it with existing architectures. The comparison was conducted using identical input image dimensions and details are given in Table~\ref{tab:performance_comparison}. Although NFDP~\cite{zixun} is compact, it showed limited robustness on complex DXA LSI images, particularly when vertebral contours were faint, merged, or affected by low signal-to-noise ratios and imaging artifacts. In contrast, VerteNet achieves a superior balance between computational efficiency and accuracy. Despite having a moderate parameter count and inference time, its multi-context feature fusion and dual-resolution attention mechanisms enable precise landmark localization across diverse imaging conditions. One can argue that, clinically, VerteNet provides more reliable landmark localization in low-contrast or artifact-prone DXA scans, ensuring consistent vertebral identification across diverse machines and patient conditions.

\begin{table}[H]
\centering
\caption{Performance comparison of VerteNet with SOTA architectures on input image size.}
\label{tab:performance_comparison}
\renewcommand{\arraystretch}{1.1}
\begin{tabular}{lccc}
\hline
\multirow{2}{*}{\textbf{Model}} & \multicolumn{3}{c}{\textbf{Performance Metrics}} \\
\cline{2-4}
 & \textbf{Parameters (M)} & \textbf{\begin{tabular}[c]{@{}c@{}}Inference\\Time (ms)\end{tabular}} & \textbf{GFLOPs} \\
\hline
GuideNet~\cite{zaid}  & 24.20 & \textbf{75.47} & 85.68 \\
HRNet~\cite{hrnet}    & 28.50 & 200.00 & 75.85 \\
NFDP~\cite{zixun}     & 12.10 & 210.00 & 27.00 \\
VerteNet (Ours)       & 24.22 & 150.27 & 76.68 \\
\hline
\end{tabular}
\end{table}

\begin{table}[]
\caption{Effect of reduction factor (r) and window size (p) in DRSA and DRCA on Model's Performance}
\label{table 4}
\begin{center}
\begin{tabular}{lllcc}
\hline
\multirow{2}{*}{\textbf{Lvl2}} & \multirow{2}{*}{\textbf{Lvl3}} & \multirow{2}{*}{\textbf{Lvl4}} & \multirow{2}{*}{\textbf{\begin{tabular}[c]{@{}c@{}}Normalized \\ Mean Error\end{tabular}}} & \multirow{2}{*}{\textbf{\begin{tabular}[c]{@{}c@{}}Normalized \\ Median Error\end{tabular}}} \\
                               &                                &                                &                                                                                            &                                                                                              \\ \hline
r=4                            & r=4                            & r=2                            & \multirow{2}{*}{5.64}                                                                      & \multirow{2}{*}{2.51}                                                                        \\
p=20                           & p=20                           & p=10                           &                                                                                            &                                                                                              \\
r=4                            & r=4                            & r=2                            & \multirow{2}{*}{5.3}                                                                       & \multirow{2}{*}{2.42}                                                                        \\
p=10                           & p=10                           & p=10                           &                                                                                            &                                                                                              \\
r=2                            & r=2                            & r=2                            & \multirow{2}{*}{5.08}                                                                      & \multirow{2}{*}{2.39}                                                                        \\
p=20                           & p=20                           & p=10                           &                                                                                            &                                                                                              \\
r=2                            & r=2                            & r=2                            & \multirow{2}{*}{4.92}                                                                      & \multirow{2}{*}{2.35}                                                                        \\
p=10                           & p=10                           & p=10                           &                                                                                            &                                                                                              \\ \hline
\end{tabular}
\vspace{-3mm}
\end{center}
\end{table}

\begin{table}[t]
\caption{Effect of Backbone on Model's Performance}
\renewcommand{\arraystretch}{1.1}
\label{table 5}
\begin{center}
\begin{tabular}{lcc}
\hline
\multirow{2}{*}{\textbf{Backbone}} & \multirow{2}{*}{\textbf{\begin{tabular}[c]{@{}c@{}}Normalized \\ Mean Error\end{tabular}}} & \multirow{2}{*}{\textbf{\begin{tabular}[c]{@{}c@{}}Normalized \\ Median Error\end{tabular}}} \\                                &   &                                                                                            \\ \hline
ResNet34 & 6.12 & 2.67 \\
EfficientNetB3 & 5.32  & 2.44  \\
EfficientNetV2S & 4.92   & 2.35  \\ \hline
\end{tabular}
\end{center}
\end{table}

\begin{table}[t]
\caption{Effect of different configurations of decoder on performance}
\label{table 6}
\begin{center}
\resizebox{\columnwidth}{!}{
\begin{tabular}{lcc}
\hline
\multirow{2}{*}{\textbf{Decoder's Configuration}}                                   & \multirow{2}{*}{\textbf{\begin{tabular}[c]{@{}c@{}}Normalized \\ Mean Error\end{tabular}}} & \multirow{2}{*}{\textbf{\begin{tabular}[c]{@{}c@{}}Normalized \\ Median Error\end{tabular}}} \\
 &  & \\ \hline
Without DRCA or DRSA  & 6.12 & 2.62 \\
With DRSA blocks only & 5.78 & 2.57 \\
With DRCA blocks only & 5.67 & 2.45 \\
\begin{tabular}[c]{@{}c@{}}With both DRCA and DRSA\\ blocks (Proposed MCFB)\end{tabular} & 4.92 & 2.35 \\ \hline
\end{tabular}}
\end{center}
\end{table}

\section{Conclusion and Future Work}
In this study, we introduced VerteNet, a deep learning architecture that incorporates a novel multi-context feature fusion block utilizing dual-resolution self- and cross-attention mechanisms. Trained on images from various DXA machines, VerteNet achieved state-of-the-art performance. VerteNet's VLL by showing its ability to estimate ROI in images to detect potential abdominal aorta cropping. This automated method ensures adequate soft tissue regions are captured to assess calcification across the entire abdominal aorta. Furthermore, as a proof of concept study, we illustrated that vertebral landmarks can generate accurate and precise IVGs, which  make it easier and quicker for readers to score AAC-24 and may help standardize the assessment between readers and improve consistency among individuals assessing AAC-24 scores on DXA LSIs. For this, although the findings are promising, the small sample size limits the generalizability, highlighting the need for further studies with larger and more diverse datasets to confirm these results. Future research could include validating this proof of concept through additional clinical experiments, examining the influence of IVGs on inter-reader variability across varying levels of expertise, or extending the proposed framework to applications like kyphosis detection and severity quantification using identified landmarks.

\vspace{3mm}
\noindent\textbf{CRediT Authorship Contribution Statement}: \textbf{Arooba Maqsood} conducted the analysis, interpreted the results, and prepared the manuscript. \textbf{Zaid Ilyas} led the deep learning framework development, co-led the analysis, and helped finalize the manuscript. Both contributed equally and share primary responsibility for the work. \textbf{Afsah Saleem} reviewed the manuscript and provided technical feedback. \textbf{John T. Schousboe, William D. Leslie, Joshua Lewis, Parminder Raina, and Jonathan M. Hodgson} contributed dataset access, clinical/epidemiological guidance, interpretation, and critical review. \textbf{Joshua Lewis} also secured funding and contributed to study design. \textbf{Syed Zulqarnain Gilani, Erchuan Zhang, and David Suter} provided technical expertise in model development and manuscript revision.

\vspace{3mm}
\noindent\textbf{Declaration of Competing Interest:} There are no competing interests to declare by the authors.

\vspace{3mm}
\noindent\textbf{Acknowledgment}: This work was supported by the Health Partners Institutional Review Board (\#A20-149), the Edith Cowan University Human Research Ethics Committee (Project Number: 20513 HODGSON), the Health Research Ethics Board at the University of Manitoba (HREB H2004:017L, HS20121), the Manitoba Health Information Privacy Committee signed consent (HIPC 2016/2017–29), and the National Health and Medical Research Council of Australia Ideas Grant (APP1183570). Additional funding was provided by the Rady Innovation Fund at the University of Manitoba, the Raine Medical Research Foundation of Australia, and the National Heart Foundation of Australia Future Leader Fellowship (ID: 102817). JRL`s salary was supported by the National Heart Foundation of Australia Future Leader Fellowship (ID: 102817). SZG's salary was partially covered by the Raine Medical Research Foundation through the Raine Priming Grant.

\vspace{3mm}
\noindent\textbf{Data Availability Statement:} The data cannot be made publicly available due to patient privacy and institutional data-sharing restrictions. The study code is publicly available at: \hyperlink{https://github.com/zaidilyas89/VerteNet}{https://github.com/zaidilyas89/VerteNet}.

\bibliographystyle{splncs04}
\bibliography{references}
\end{document}